# A Polynomial-Time Algorithm for Deciding Markov Equivalence of Directed Cyclic Graphical Models

**Thomas Richardson**
Philosophy Department
Carnegie-Mellon University
Pittsburgh, PA 15213
e-mail: tsr+@andrew.cmu.edu

## Abstract

Although the concept of d-separation was originally defined for directed *acyclic* graphs (see Pearl 1988), there is a natural extension of the concept to directed *cyclic* graphs. When exactly the same set of d-separation relations hold in two directed graphs, no matter whether respectively cyclic or acyclic, we say that they are Markov equivalent. In other words, when two directed cyclic graphs are Markov equivalent, the set of distributions that satisfy a natural extension of the Global Directed Markov Condition (Lauritzen et al. 1990) is exactly the same for each graph. There is an obvious exponential (in the number of vertices) time algorithm for deciding Markov equivalence of two directed cyclic graphs; simply check all of the d-separation relations in each graph. In this paper I state a theorem that gives necessary and sufficient conditions for the Markov equivalence of two directed cyclic graphs, where each of the conditions can be checked in polynomial time. Hence, the theorem can be easily adapted into a polynomial time algorithm for deciding the Markov equivalence of two directed cyclic graphs. Although space prohibits inclusion of correctness proofs, they are fully described in Richardson (1994b).

## 1. INTRODUCTION

Directed Cyclic Graphical Models (DCGs), described in Spirtes (1995), are a generalization of DAG models (Pearl 1988). Spirtes has shown that linear simultaneous structural equation models, which are widely used to represent feedback in engineering and the social sciences, satisfy a natural extension of the Global Directed Markov Property for cyclic graphs. (See also Koster 1994). Pearl (1993) has also investigated rules for predicting the effects of interventions in simultaneous equation models.

Markov equivalence for DAGs was characterized by Verma and Pearl (1990, 1992) and for more general chain graphs by Frydenberg (1990). The problem of characterizating Markov equivalence for graphs with cycles was posed (independently) by Koster (1994) for "reciprocal graphs" (a generalization of chain graphs).[1] Similar questions were raised by Basmann (1965), Stetzl (1986) and Lee (1987).

A greater understanding of the relationship between cyclic causal systems and statistical independencies will facilitate the construction of efficient discovery algorithms; these algorithms will output the class of Directed Cyclic Graphical models compatible with data given as input, in situations where the underlying causal structure contains loops. This work also provides a principled basis for studying relations between cyclic graphs and time series. (See Spirtes *et al.* 1993, Richardson 1994b, Sobel 1994)

## 2. DAG MODELS

A Directed Graph $G$ consists of an ordered pair $<V,E>$, where $V$ is a set of vertices, and $E$ is a set of directed edges between vertices.[2] If there are no directed cycles[3] in $E$, then $<V,E>$ is called a Directed Acyclic Graph or (DAG). A DAG *model* is an ordered pair $<G, P>$, consisting of a DAG $G$, and a joint probability distribution $P$, over the set $V$, in which certain conditional independence relations, encoded by the graph, are true.[4] The independencies encoded by a given graph are determined by a graphical criterion called d-separation, as explained in Pearl (1988). The following definition can be applied to cyclic and acyclic cases, and is equivalent to Pearl's in the latter. We first require the following definition:

---

[1] For a definition of Chain Graph see Whittaker(1990) pp.77-79.
[2] If $<A,B> \in E$, A, B distinct, then we say that there is an edge *from* A *to* B, and we represent this as A→B. If $<A,B> \in E$ *or* $<B,A> \in E$, then in either case we say that there is an edge *between* A and B. There can be at most one edge $<A,B> \in E$, (since E is a set), though it is possible to have $<A,B>$ *and* $<B,A> \in E$.
[3] By a 'directed cycle' we mean a directed path $X_0 \to X_1 ... \to X_{n-1} \to X_0$ of n distinct vertices, where n≥2. A directed graph is *acyclic* if it contains no directed cycles.
[4] Since the elements of V are both vertices in a graph, and random variables in a joint probability distribution, we shall use the terms 'variable' and 'vertex' interchangeably.

**Definition:** Child, Parent, Descendant, Ancestor

If there is an arrow from A to B (A→B), then we say that A is a parent of B, and B is a child of A. We define the 'descendant' relation as the transitive reflexive closure of 'child', and similarly, 'ancestor' as the transitive reflexive closure of 'parent', so every vertex is its own ancestor and descendant.

**Definition:** d-connection /d-separation for directed graphs

For disjoint sets of vertices, **X**, **Y** and **Z**, **X** is *d-connected to* **Y** *given* **Z** if and only if for some X∈ **X**, and Y∈ **Y**,[5] there is an (acyclic) undirected path **U** between X and Y, such that:

(i) If there is an edge between A and B on **U**, and an edge between B and C on **U**, and B∈ **Z**, then B is a collider between A and C relative to **U**, i.e. A→B←C on the path **U**.

(ii) If B is a collider between A and C relative to **U**, then there is a descendant D, of B, and D∈ **Z**.

For disjoint sets of vertices, **X**, **Y** and **Z**, if **X** and **Y** are not d-connected given **Z** then **X** and **Y** are said to be *d-separated* given **Z**.

### 2.1. THE GLOBAL DIRECTED MARKOV CONDITION

In a DAG model $<G,P>$ the following constraint relates $G$ and $P$:

A DAG model $<G,P>$ is said to satisfy the Global Directed Markov Property whenever for all disjoint sets of variables **A**, **B** and **C**, **A** is independent of **B** given **C** in $P$ if **A** is d-separated from **B** given **C** in $G$.

This condition is of great theoretical importance since a wide range of statistical models can be represented as DAG models satisfying the Global Directed Markov Condition, including recursive linear structural equation models with independent errors, regression models, factor analytic models, path models, and discrete latent variable models (via appropriate extensions of the formalism.) An alternative, though equivalent, definition of the Global Directed Markov Property is given by Lauritzen *et al.* (1990).

We introduce the following notion of Markov equivalence:

**Markov Equivalence for Graphs (Cyclic or Acyclic)**

Graphs $G_1$ and $G_2$ are *Markov equivalent* if every distribution which satisfies the Global Directed Markov condition with respect to one graph satisfies it with respect to the other, and vice versa.

Since the Global Directed Markov condition only places conditional independence constraints on distributions, under this definition, two graphs are Markov equivalent if and only if the same d-separation relations hold in both graphs. In fact, a result of Spirtes(1994) shows that d-separation is complete for directed (cyclic or acyclic) graphs, hence, if two graphs are Markov equivalent as defined above, then they entail the same conditional independencies.

### 2.2 THE LOCAL DIRECTED MARKOV CONDITION

For acyclic graphs, the Global Directed Markov Condition is equivalent to another condition:

A DAG model $<G,P>$ is said to satisfy the Local Directed Markov Property if every variable A in $G$ is independent of all variables other than its parents and descendants in $G$, given its parents in $G$.

For acyclic graphs the Global and Local Directed Markov Conditions are equivalent (Lauritzen *et al.* 1990). Hence, a characterization of when two acyclic graphs are equivalent under the Global Directed Markov Condition is also a characterization of equivalence under the Local condition.

### 2.3 DIRECTED CYCLIC GRAPHS

The Global Directed Markov Condition was originally defined for *acyclic* graphs. However, the question naturally arises as to whether the condition can be applied to cyclic graphs. Given a careful definition of the notion of (undirected) path, to allow for the fact that there may be more than one edge between a given pair of variables,[6] the definition can be applied directly. The same is also true for the Local Directed Markov Condition. In cyclic graphs, the natural extensions of the Local and Global Directed Markov Conditions[7] are no longer equivalent, as the following graph (from Whittaker 1990) shows:

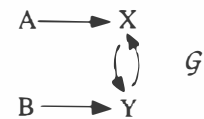

Figure 1: A Directed Cyclic Graph

Under the Local Directed Markov Condition B is independent of X given A and Y. Yet under the Global Directed Markov Condition this is not so since the path X→A←B d-connects X and B given A and Y.[8]

One might wonder whether or not the conditional independencies entailed by the Global Directed Markov condition applied to a Directed Cyclic Graph correspond to a natural class of statistical models. In fact Spirtes (1995) has shown that the conditional independencies which hold in non-recursive linear structural equation models[9] are precisely those entailed by the Global Directed Markov condition, applied to the graph naturally associated with a structural equation model[10] with independent errors.

---

[5] We use bold face upper case Roman letters (**V**) to denote sets of variables, and plain face Roman letters (V) to denote single variables.

[6] Thus a sequence of vertices does not necessarily define a unique path.

[7] We shall refer to these extensions as the Local and Global Directed Markov Conditions since the definitions carry over from the acyclic case without change.

[8] It is easy to prove that the conditional independencies which hold in the graph in Fig. 1 under the Global Directed Markov Condition cannot be represented by any Chain Graph.

Non-recursive structural equation models are used in a wide variety of fields to model linear systems in which feedback is present:

In **economics**, non-recursive linear structural equation models are used in price theory: the price of a good in a market may be dependent on the quantity either demanded or supplied, while these quantities themselves may be influenced by the (expectation of) price that suppliers may have.

In **biology**, these models are used to model systems that act to maintain 'dynamic equilibria', of which there are many instances: from the molecular processes that control the enzymatic production of chemicals, to the predator-prey relationships which curb population growth.

Models of this kind are also exploited in fields as diverse as sociology, robotics and psychology, where some types of neural net are of this form.

## 3 MARKOV EQUIVALENCE FOR DIRECTED GRAPHS

The DAG formalism has had fruitful results in many areas: there is now a relatively clear causal interpretation of these models, there are efficient procedures for determining the statistical indistinguishability of DAG's, asymptotically reliable algorithms for generating a class of DAG models from sample data and background knowledge, etc. A crucial element in these investigations was a 'local' characterization of Markov equivalence. This local characterization was essential in allowing the construction of efficient algorithms which could search the whole class of DAG models to find those which fitted the given data under certain assumptions (See Spirtes *et al.* 1993).

### 3.1 MARKOV EQUIVALENCE FOR ACYCLIC GRAPHS

In the acyclic case there is a relatively simple characterization of the Markov equivalence class that leads directly to an $O(n^3)$ algorithm. We first require the following:

**Definition:** Unshielded Collider and Non-Collider[11]

In a directed graph $G$, the triple $<A,B,C>$ forms an *unshielded collider* in $G$, if there is no edge between A and C (neither $A \to C$ nor $C \to A$), but there are edges from A to B, and from C to B, i.e. $A \to B \leftarrow C$.

If there is no edge between A and C, there is an edge between A and B, and an edge between B and C, but $<A,B,C>$ is not an unshielded collider, then we say it is an *unshielded non-collider*, i.e. $A \to B \to C$, $A \leftarrow B \to C$, or $A \leftarrow B \leftarrow C$.

**Equivalence Theorem for Acyclic Graphs** (Verma and Pearl 1990, Frydenberg 1990)

Two DAGs, $G_1$, $G_2$, are Markov equivalent if and only if

**(a)** $G_1$ and $G_2$ contain the same vertices

**(b)** There is an edge between A and B in $G_1$ if and only if there is an edge between A and B in $G_2$

**(c)** $G_1$ and $G_2$ have the same unshielded colliders

Conditions (a), (b) and (c) imply (logically) a fourth condition:

**(d)** $G_1$ and $G_2$ have the same unshielded non-colliders

Below we show three examples of acyclic Markov equivalence classes:

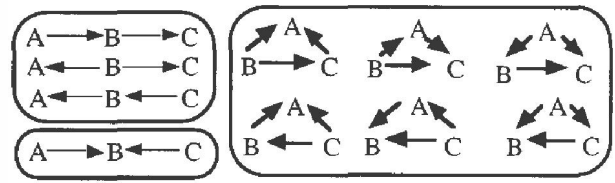

Figure 2: Three Acyclic Markov Equivalence Classes

Conditions (a), (b) and (c) above lead to an $O(n^3)$ algorithm for checking the Markov equivalence of two acyclic graphs on n variables; this follows from the fact that (b) mentions pairs of variables, while (c) mentions triples. Although d-separation allows us to check any given conditional independence, there are $O(2^n)$-many conditional independencies,[12] thus d-separation alone does not provide a feasible test for Markov equivalence.

### 3.2 MARKOV EQUIVALENCE IN THE CYCLIC CASE

This raises the question of whether conditions similar to (a), (b) and (c) exist for the cyclic case. The answer is that such a set of conditions do exist. The conditions are considerably more complicated, but still lead to a polynomial algorithm, though of $O(n^9)$ or $O(n^3 e^4)$ where e is the number of edges in the graph.[13] (Richardson, 1994b). Using this result we can show that certain sets of d-separation relations hold in no acyclic graph, but do hold in certain cyclic graphs. It also provides a first step towards a discovery algorithm which will construct models from conditional independencies present in data; the output of such a discovery algorithm is a Markov equivalence class of models. Characterizing Markov equivalence requires noticing important differences between properties of d-separation in acyclic and in cyclic graphs, and characterizing properties peculiar to the latter.

---

[9] A non-recursive structural equation model is one in which the matrix of coefficients is not in lower triangular form, for any ordering of the equations. (Bollen 1989)

[10] i.e. the directed graph in which X is a parent of Y, if and only if the coefficient of X in the structural equation for Y is not fixed at zero by the model.

[11] 'Colliders' correspond to 'head-to-head' nodes in the terminology of Verma & Pearl (1988).

[12] This follows from the assumption of Faithfulness, without which there would be $o(2^{n^3})$ many conditional independencies to check.

[13] It should be stressed that this is a (loose) worst case complexity bound, the expected case may be much lower.

Condition (a) is obviously necessary for equivalence in the cyclic case - if two graphs contain different sets of variables, then trivially there will be different d-connection relations which hold in these graphs.

### 3.3 REAL & VIRTUAL ADJACENCIES

In the cyclic case the condition (b) (§3.1) requiring graphs to have edges between the same vertices is no longer necessary for Markov equivalence.[14] This can be seen by considering the following two Markov equivalent models:

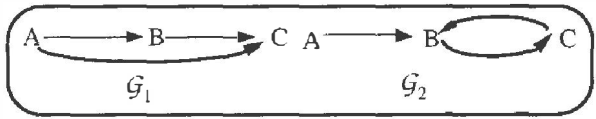

Figure 3: Markov Equivalent DCGs with different adjacencies

No d-separation relations hold in either structure. In $G_1$ there is an edge between A and C, but there is no edge between A and C in $G_2$. We introduce the following definitions:

**Definition:** Virtually Adjacent

A and C are said to be *virtually adjacent* in cyclic $G$ if and only if A and C have a common child B, such that B is an ancestor of A or of C.[15]

**Definition:** Really Adjacent

We incorporate edges into the notion of adjacency by saying that if there is an edge between A and C (A→C or A←C), then A and C are *really adjacent*.[16]

Thus in Figure 3, in $G_1$, A and C are really adjacent, while in $G_2$ A and C are virtually adjacent. Virtual adjacencies can only occur in cyclic graphs (since B is in a cycle with A or C). Condition (b) is necessary for Markov equivalence in the acyclic case since in an acyclic graph A and B are d-connected given every subset of the other variables if and only if there is an edge between A and B, i.e. if A and B are really adjacent. In the cyclic case A and B are d-connected given every subset of the other variables if and only if A and B are either really *or* virtually adjacent. It follows that given two Markov equivalent cyclic graphs $G_1$, $G_2$, the following is true:

**(1)** If A and B are either virtually or really adjacent in $G_1$, then A and B are either virtually or really adjacent in $G_2$.

In fact, if a cyclic graph contains a virtual adjacency then there is always a Markov equivalent cyclic graph in which that adjacency is real. In the context of directed cyclic graphs we will use the term 'adjacency' to mean 'real or virtual adjacency'.

### 3.4 UNSHIELDED CONDUCTORS

In the cyclic case, condition (d) (§3.1) requiring the same unshielded non-colliders is no longer necessary for Markov equivalence. This can be seen from the following Markov equivalent graphs $G_1$ and $G_2$, in each of which A and C are d-separated given {B,D} (and no other set).

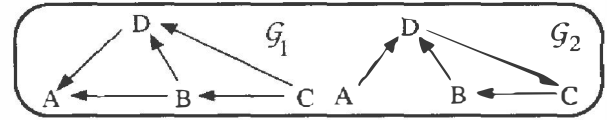

Figure 4: Markov Equivalent DCGs With Different Unshielded Non-Colliders

<A,B,C> forms an unshielded non-collider in $G_1$, but not in $G_2$. This leads us to make the following definition:

**Definition:** Unshielded Conductor and Unshielded Non-Conductor

In a cyclic graph $G$, we say triple of vertices <A,B,C> forms an *unshielded conductor* if:

(i) A and B are adjacent, B and C are adjacent, A and C are not adjacent

(ii) B is an ancestor of A or C

If <A,B,C> satisfies (i), but B is not an ancestor of A or C, we say <A,B,C> is an *unshielded non-conductor*.

In the acyclic case condition (d) was necessary for Markov equivalence, since there is a set of d-connection and d-separation relations among the vertices A, B, and C which hold in an acyclic graph if and only if <A,B,C> is an unshielded non-collider. In the cyclic case the same set of d-connection and d-separation relations hold in the graph if and only if <A,B,C> is an unshielded conductor.[17] Thus if cyclic graphs $G_1$, $G_2$ are Markov equivalent then:

**(2)** <A,B,C> is an unshielded conductor in $G_1$ if and only if <A,B,C> is an unshielded conductor in $G_2$

Thus, in terms of d-separation relations, an unshielded conductor is the cyclic analogue of the unshielded non-collider; in an acyclic graph every unshielded conductor will be an unshielded non-collider.

### 3.5 UNSHIELDED PERFECT NON-CONDUCTORS

Condition (c) of §3.1 requiring that two graphs have the same unshielded colliders is also no longer a necessary condition for Markov equivalence. This can be seen by considering the triple <A,D,C> in Fig. 5. $G_1$ and $G_2$ are Markov equivalent; in both graphs A and C are d-separated given ∅, and only this set. <A,D,C> is an unshielded collider in $G_1$, but this is not the case in $G_2$.

---

[14] A similar point is made in Whittaker (1990).

[15] Note that every vertex is its own descendant and ancestor.

[16] The terms 'Parent', 'Child', 'Ancestor' and 'Descendant' continue to refer exclusively to real edges – virtual adjacencies are not oriented.

[17] The set of d-separation and d-connetion relations is as follows: A and B are d-connected given every subset of the other variables. B and C are also d-connected given every subset of the other variables. A and C are d-connected given every subset of the other variables that excludes B.

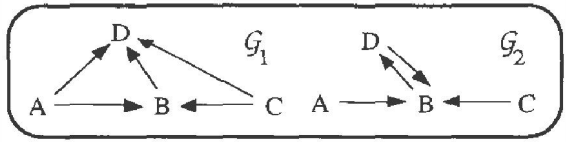

Figure 5: Markov Equivalent DCGs With Different Unshielded Colliders

Consideration of this example and others similar to it leads us to make the following definition:

**Definition:** Unshielded Perfect and Imperfect Non-Conductors

In a cyclic graph $G$, we say triple of vertices <A,B,C> is an *unshielded perfect non-conductor* if:

(i) A and B are adjacent, B and C are adjacent, but A and C are not adjacent.

(ii) B is not an ancestor of A or C.

(iii) B is a descendant of a common child of A and C.

If <A,B,C> satisfies (i) and (ii) but B is not a descendant of a common child of A and C, we say <A,B,C> is an *unshielded imperfect non-conductor*.

As in the previous cases condition (c) of §3.1 is necessary for Markov equivalence in the acyclic case because there is a set of d-connection and d-separation relations among the vertices A, B, and C which hold in an acyclic graph if and only if <A,B,C> is an unshielded collider. In a cyclic graph the same set of d-connection and d-separation relations hold if and only if <A,B,C> is an unshielded perfect non-conductor.[18] Hence the following is also necessary for Markov equivalence:

(3) <A,B,C> is an unshielded perfect non-conductor in $G_1$ iff it is also an unshielded perfect non-conductor in $G_2$.

Thus, in terms of d-connection relations, unshielded perfect non-conductors are the cyclic analogue to unshielded colliders in the acyclic case. However, in an acyclic graph every unshielded triple is either an unshielded collider or non-collider, whereas it is not the case that in a cyclic graph every unshielded triple is either an unshielded conductor or an unshielded perfect non-conductor. A triple may form an unshielded *imperfect* non-conductor, in the case where the following conditions hold:

(i) A and B are adjacent, and B and C are adjacent, but A and C are not adjacent.

(ii) B is not an ancestor of A or C.

(iii) B is *not* a descendant of a common child of A and C

Conditions (i) and (ii) imply that <A,B,C> is an unshielded non-conductor, but since, by condition (iii) B is not a descendant of a common child of A and C, <A,B,C> is not an unshielded perfect non-conductor.

We have just seen that unshielded conductors, and unshielded perfect non-conductors can be uniquely characterized by d-connection relations. It follows from this that the d-connection relations that hold among a triple which forms an unshielded imperfect non-conductor do not hold in any acyclic graph (even with latent variables). This provides a criterion for detecting the presence of feedback.[19]

### 3.6 CONTRAST WITH THE ACYCLIC CASE: NON-LOCALITY

In the acyclic case, if two graphs are *not* Markov equivalent then there will be two vertices A, B at most two edges apart[20] in one of the graphs, such that for some subset **R** of the other vertices, A and B are d-separated given **R** in one graph, and d-connected given **R** in the other. This means that in the acyclic case we need only look at the structure of triples of really adjacent vertices in order to establish that two graphs are Markov equivalent. This is not true for the cyclic case, as the following two graphs which are not Markov equivalent show:

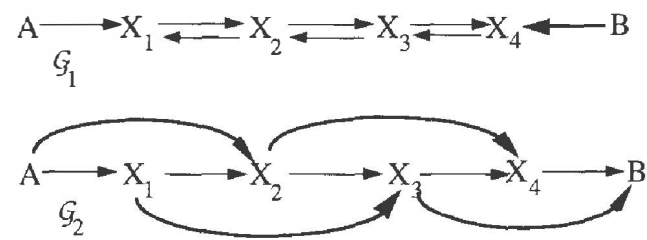

Figure 6: Two Non-Equivalent DCGs That Are Locally Equivalent

Every d-separation relation which holds in $G_2$, also holds in $G_1$. However, in $G_1$, A and B are d-separated given $\varnothing$ while A and B are d-connected given $\varnothing$ in $G_2$, and this is the only d-separation relation which holds in $G_1$ and not in $G_2$. Moreover A and B are more than two edges apart in both graphs, and clearly these graphs could be extended by increasing the number of X's so that A and B were arbitrarily many edges apart.

This is why the cyclic equivalence algorithm is of higher (though still polynomial) complexity; cyclic graphs cannot be compared by checking that all 'local' subgraphs are Markov equivalent. $G_1$ and $G_2$ also show that the conditions (1)–(3), though necessary, are not sufficient for Markov equivalence since all three conditions are satisfied by these graphs which are not Markov equivalent. The full set of necessary and sufficient conditions is given in the Cyclic Equivalence Theorem stated in the Appendix.

---

[18]The set of d-separation and d-connection relations is as follows: A and B are d-connected given every subset of the other vertices. B and C are also d-connected given every subset of the other vertices. A and C are d-connected given every subset of the other vertices that includes B.

[19]See Proposition 2 in the Appendix.
[20]i.e. the shortest undirected path from A to B contains at most one other vertex.

### 3.6 The Orientation of Cycles

The Cyclic Equivalence Theorem,[21] has the following interesting consequence:

Given a graph $G$ with a cycle $C$, there is a Markov equivalent graph $G^*$, in which $C$ is replaced by another cycle $C^*$, having the opposite orientation to $C$. Thus if $C$ is clockwise, $C^*$ is anti-clockwise, and vice-versa.

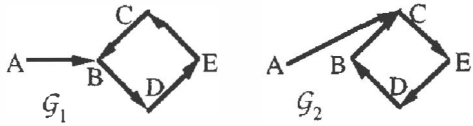

Figure 7: Two Markov Equivalent DCGs With Contra-Oriented Directed Cycles.

In $G_1$, the cycle <C,B,D,E> has anti-clockwise orientation, while in the Markov equivalent graph $G_2$, the corresponding cycle <C,E,D,B> has clockwise orientation. One important consequence is that it is impossible to orient a cycle merely using conditional independence information.

## 4 THE EQUIVALENCE ALGORITHM

The set of conditions given in the Appendix lead directly to a polynomial time algorithm for determining the Markov equivalence of two cyclic graphs. All that is required is that, for each graph, we compile a list of the features mentioned in conditions (0)-(6) in the Equivalence Theorem, and compare the results.

It will be obvious that most of these conditions require only 'local' information about the graph. The only conditions for which this is less obvious are those mentioning mutually exclusive (m.e.) unshielded conductors[22] (conditions (4) and (6)), since such pairs of unshielded conductors may be arbitrarily far apart in the graph.[23] However, in fact these conditions can be checked in polynomial time: Condition (4) only requires us to check that whenever a special kind of path (an uncovered itinerary) exists in one graph between two unshielded conductors, that there is *some* path (not necessarily the same one) of the same kind which exists between these conductors in the other graph. This can be achieved in polynomial time ($O(n^6)$) using a suitably adapted version of the Path Sum Algorithm. (Aho, Hopcroft & Ullman (1974)). Condition (6), which also refers to m.e. unshielded conductors can easily be checked once we have calculated the pairs of m.e. unshielded conductors for each graph, since it only requires us to check whether one vertex is an ancestor of another.

## 5 APPENDIX

### 5.1 BASIC GRAPHICAL CONCEPTS

**Definition:** Edge

An arrow from A to B (A→B) is called an edge *from* A *to* B. An arrow from A to B (A→B) or from B to A (B→A) are both called an *edge between* A and B.

**Definition:** Directed Path

A sequence of distinct edges <$E_1,...,E_n$> in $G$ is a *directed path* from $V_1$ to $V_{n+1}$ if and only if there exists a sequences of vertices <$V_1,...,V_{n+1}$> such that for $1 \leq i \leq n$ <$V_i,V_{i+1}$> = $E_i$, i.e. $V_1 \to V_2 \to ... \to V_{n+1}$. The path is called *acyclic* if the vertices <$V_1,...,V_{n+1}$> are distinct.

**Definition:** Undirected Path

A sequence of distinct edges <$E_1,...,E_n$> in $G$ is an *undirected path* if and only if there exists a sequences of vertices <$V_1,...V_{n+1}$> s.t. for $1 \leq i \leq n$ either <$V_{i+1},V_i$> = $E_i$ or <$V_i,V_{i+1}$> = $E_i$. The path is called *acyclic* if the vertices <$V_1,...,V_{n+1}$> are distinct.

**Definition:** Collider (Non-Collider) relative to edges or a path.

Given three vertices A, B and C such that there is an edge between A and B, and between B and C, then if the edges 'collide' at B, we say B is a *collider* between A and C, *relative to these edges* i.e. A→B← C.

Otherwise, if there is an edge between A and B, and between B and C, but the edges do not 'collide', we will say that B is a *non-collider* between A and C, *relative to these edges*. i.e. A is a non-collider: A→B→C, A←B→C, A←-B←-C.

### 5.2 CYCLIC EQUIVALENCE THEOREM

We give below the Equivalence Theorem in full. However, this requires three more definitions:

**Definition:** Itinerary

If <$X_0,X_1,...X_{n+1}$> is a sequence of distinct vertices s.t. $\forall i\ 0 \leq i \leq n$, $X_i$ and $X_{i+1}$ are really or virtually adjacent then we will refer to <$X_0,X_1,...X_{n+1}$> as an *itinerary*.

**Definition:** Mutually Exclusive Unshielded Conductors (with respect to an itinerary)

If <$X_0,...X_{n+1}$> is an itinerary such that:

(i) $\forall t\ 1 \leq t \leq n$, <$X_{t-1},X_t,X_{t+1}$> is an unshielded conductor.

(ii) $\forall k\ 1 \leq k \leq n$, $X_{k-1}$ is an ancestor of $X_k$, and $X_{k+1}$ is an ancestor of $X_k$.

(iii) $X_0$ is *not* a descendant of $X_1$, and $X_n$ is *not* an ancestor of $X_{n+1}$, then <$X_0,X_1,X_2$> and <$X_{n-1},X_n,X_{n+1}$> are *mutually exclusive (m.e.) unshielded conductors on the itinerary* <$X_0,...X_{n+1}$>.

---

[21]See the Appendix.
[22]See the Appendix for the definition of mutually exclusive unshielded conductors.
[23]This is essentially the point made in §3.5. See the Appendix.

**Definition:** Uncovered itinerary

If $<X_0,...X_{n+1}>$ is an itinerary such that $\forall i,j\ 0 \leq i < j-1 < j \leq n+1$ $X_i$ and $X_j$ are not adjacent in the graph then we say that $<X_0,...X_{n+1}>$ is an *uncovered itinerary*. i.e. an itinerary is uncovered if the only vertices on the itinerary which are adjacent to other vertices on the itinerary, are those that occur consecutively on the itinerary.

**Cyclic Equivalence Theorem (Richardson 1994b)**

Cyclic graphs $G_1$ and $G_2$ are Markov equivalent if and only if the following conditions hold:

(0) $G_1$ and $G_2$ contain the same vertices.

(1) $G_1$ and $G_2$ have the same adjacencies.

(2) $G_1$ and $G_2$ have the same unshielded conductors.

(3) $G_1$ and $G_2$ have the same unshielded perfect non-conductors.

(4) For any pair of unshielded conductors $<A,B,C>$ and $<X,Y,Z>$ (in both $G_1$ and $G_2$), $<A,B,C>$ and $<X,Y,Z>$ are m.e. unshielded conductors on some uncovered itinerary $\mathbf{P} \equiv <A,B,C,...X,Y,Z>$ in $G_1$ if and only if $<A,B,C>$ and $<X,Y,Z>$ are m.e. unshielded conductors on some uncovered itinerary $\mathbf{Q} \equiv <A,B,C,...X,Y,Z>$ in $G_2$.[24]

(5) If $<A,X,B>$, and $<A,Y,B>$ are unshielded imperfect non-conductors, then X is an ancestor of Y in $G_1$ iff X is an ancestor of Y in $G_2$.[25]

(6) If $<A,B,C>$ and $<X,Y,Z>$ are m.e. unshielded conductors on some uncovered itinerary $\mathbf{P} \equiv <A,B,C,...X,Y,Z>$ (in both $G_1$ and $G_2$), and $<A,M,Z>$ is an imperfect non-conductor (in both $G_1$ and $G_2$), then M is a descendant of B in $G_1$ if and only if M is a descendant of B in $G_2$.

In Richardson (1994b), Ch.3 §4 I show that conditions (0)-(6) are logically independent.

**Proposition 1** (Richardson 1994b)

Given a directed cyclic graph $G$, there is a directed acyclic graph $G^*$ that is Markov equivalent to $G$ if and only if:

(1) there is no triple $<X,Y,Z>$ s.t. $<X,Y,Z>$ forms an unshielded imperfect non-conductor in $G$, and

(2) there are no pairs of triples $<A,B,C>$ and $<X,Y,Z>$ in $G$ such that $<A,B,C>$ and $<X,Y,Z>$ are m.e. unshielded conductors on some uncovered itinerary $\mathbf{P} \equiv <A,B,C,...X,Y,Z>$ in $G$.

### 5.4 CYCLIC CLASSIFICATION ALGORITHM

Below I give an algorithm that will output the list of features given in the Cyclic Equivalence Theorem, given a directed (cyclic or acyclic) graph $G$ with vertex set $\mathbf{V}$ as input. To determine whether two graphs $G_1$ and $G_2$ are equivalent all that is required is to compare the lists of features and undirected adjacency graphs ($\mathcal{H}_{Adj}$) produced by this procedure. (Note that for Markov equivalent graphs the graphs $G_{An}$ will not in general be the same, and so should not be compared.)

**Input:** A directed graph $G$

**Output:** A list of the features and an undirected adjacency graph $\mathcal{H}_{Adj}$ given in the cyclic equivalence theorem.

A) Form the directed graph $G_{An}$, which contains an edge $A \rightarrow B$ if and only if there is a directed path from A to B, $A \rightarrow ... B$, in $G$. This can be achieved by applying the Path Sum Algorithm (Aho, Hopcroft & Ullman, 1974) to $G$.

B) Form the undirected graph, $\mathcal{H}_{Adj}$, as follows:

Put an edge A—B in $\mathcal{H}_{Adj}$ if and only if:

(1) there is an edge $A \rightarrow B$ or $B \rightarrow A$ in $G$, or

(2) there is a vertex C such that $A \rightarrow C \leftarrow B$ in $G$ and either $C \rightarrow B$, or $C \rightarrow A$ in $G_{An}$.

Thus there is an edge A—B in $\mathcal{H}_{Adj}$ if and only if A and B are really or virtually adjacent in $G$.

C) Form a set of ordered triples of vertices, **unshielded-conductors**:

Put $<A,B,C>$ into **unshielded-conductors** if and only if:

(1) A,B,C are distinct,

(2) A—B, B—C, in $\mathcal{H}_{Adj}$, but there is no edge A—C in $\mathcal{H}_{Adj}$, and

(3) either $B \rightarrow A$, or $B \rightarrow C$, (or both) in $G_{An}$.

D) Form a set of ordered triples of vertices, **unshielded-perfect-non-conductors:** Put $<A,B,C>$ into **unshielded-perfect-non-conductors** if and only if:

(1) A,B,C are distinct,

(2) A—B, B—C, in $\mathcal{H}_{Adj}$, but there is no edge A—C in $\mathcal{H}_{Adj}$,

(3) $<A,B,C> \notin$ **unshielded-conductors**, and

(4) either there is a vertex D such that $A \rightarrow D \leftarrow C$ in $G$ and $D \rightarrow B$ in $G_{An}$, or $A \rightarrow B \leftarrow C$ in $G$.

E) Form a set of ordered sextuples of vertices, **m.e.-conductors:**

For each sextuple $<A,B,C,D,E,F>$ satisfying:

(1) $C \neq F$, and $B \neq E$

(2) $<A,B,C>, <D,E,F> \in$ **unshielded-conductors**

(3) the edges $B \rightarrow A$, and $E \rightarrow F$ are not in $G_{An}$.

(4) the edge A—F is not in $\mathcal{H}_{Adj}$, and

(5) either the edge B—E is not in $\mathcal{H}_{Adj}$, or B=D and C=E,

form an undirected subgraph $\mathcal{K}$ of $\mathcal{H}_{Adj}$, as follows:

Initially let $\mathcal{K} = \mathcal{H}_{Adj}$.

Remove from $\mathcal{K}$ all vertices adjacent to A,B,E or F, except for B,C, D and E. Remove all edges X—Y from $\mathcal{K}$ unless $X \rightarrow Y$ in $G_{An}$, and $X \leftarrow Y$ in $G_{An}$.

---

[24] i.e. $\mathbf{P}$ and $\mathbf{Q}$ need not be the same path, though the first and last three vertices of each path are the same.

[25] It is a logical consequence of Conditions(1)-(3) that $G_1$ and $G_2$ have the same unshielded imperfect non-conductors.

Next remove A and F from $\mathcal{K}$. Now apply the Path Sum Algorithm to $\mathcal{K}$ in order to determine whether there is a path from B to E. If there is a path from B to E then put <A,B,C,D,E,F> in **m.e.-conductors**.

**F)** Form a set of ordered pairs <W,V>, **imperfect-non-conductor-ancestors**, as follows:

If <W,V> are s.t. there exist vertices A and F satisfying:

(1) A–W, W–F, A–V, V–F in $\mathcal{H}_{Adj}$, but there is no edge A–F in $\mathcal{H}_{Adj}$

(2) <A,W,F> and <A,V,F> ∉ **unshielded-conductors**

(3) <A,W,F> and <A,V,F> ∉ **unshielded-perfect-non-conductors**

(4) V→W in $\mathcal{G}_{An}$.

then put <W,V> into **imperfect-non-conductor-ancestors**.

**G)** Form a set of ordered pairs <W,V>, **m.e.-conductors-to-imperfect-non-conductor-ancestors**:

If <W,V> are s.t. there exist vertices A,C,D,E,F satisfying:

(1) A–W, W–F in $\mathcal{H}_{Adj}$, but there is no edge A–F in $\mathcal{H}_{Adj}$

(2) <A,W,F> ∉ **unshielded-conductors**

(3) <A,W,F> ∉ **unshielded-perfect-non-conductors**

(4) there is a sextuple <A,V,C,D,E,F>∈ **m.e.-conductors**

(5) V→W in $\mathcal{G}_{An}$

then put <W,V> into **m.e.-conductors-to-imperfect-non-conductor-ancestors**.

### Acknowledgements
I thank P. Spirtes, C. Glymour, R. Scheines & C. Meek for helpful conversations. Research for this paper was supported by ONR grant N00014-93-1-0568.